\documentclass[sigconf]{acmart}
\usepackage{multirow}
\usepackage{url}
\usepackage{array}
\usepackage{subfigure}
\usepackage{balance}
\usepackage[misc]{ifsym}
\usepackage{graphicx}
\usepackage{enumitem}
\usepackage{makecell}
\AtBeginDocument{%
  \providecommand\BibTeX{{%
    \normalfont B\kern-0.5em{\scshape i\kern-0.25em b}\kern-0.8em\TeX}}}

\setcopyright{acmcopyright}
\copyrightyear{2022}
\acmYear{2022}
\setcopyright{acmcopyright}
\acmConference[WSDM '22] {Proceedings of the Fifteenth ACM International Conference on Web Search and Data Mining}{February 21--25, 2022}{Tempe, AZ, USA.}
\acmBooktitle{Proceedings of the Fifteenth ACM International Conference on Web Search and Data Mining (WSDM '22), February 21--25, 2022, Tempe, AZ, USA}
\acmPrice{15.00}
\acmDOI{10.1145/3488560.3498405}
\acmISBN{978-1-4503-9132-0/22/02}

\begin{document}

\fancyhead{}
\title{Knowledge Enhanced Sports Game Summarization}

\author{
Jiaan Wang$^{1*}$, Zhixu Li$^{2*(\textrm{\Letter})}$, Tingyi Zhang$^{1}$, Duo Zheng$^{3}$, Jianfeng Qu$^{1(\textrm{\Letter})}$ \\ An Liu$^{1}$, Lei Zhao$^{1}$, and Zhigang Chen$^{4}$
}
\makeatletter
\def\authornotetext#1{
 \g@addto@macro\@authornotes{%
 \stepcounter{footnote}\footnotetext{#1}}%
}
\makeatother

\authornotetext{The first two authors made equal contributions to this work.}

\affiliation{%
  \institution{$^{1}$ School of Computer Science and Technology, Soochow University, Suzhou, China}
  \country{}
}
\affiliation{%
  \institution{$^{2}$ Shanghai Key Laboratory of Data Science, School of Computer Science, Fudan University}
  \country{}
}
\affiliation{%
  \institution{$^{3}$ Beijing University of Posts and Telecommunications \quad $^{4}$ iFLYTEK Research, Suzhou}
  \country{}
}

% \affiliation{%
%   \institution{$^{4}$ iFLYTEK Research, Suzhou}
%   \country{}
% }

% \email{jawang1@stu.suda.edu.cn,zhixuli@fudan.edu.cn, jfqu@suda.edu.cn}

\email{{jawang1, tyzhang1}@stu.suda.edu.cn,zhixuli@fudan.edu.cn, zd@bupt.edu.cn}
\email{{jfqu, anliu, zhaol}@suda.edu.cn, zgchen@iflytek.com}

\begin{abstract}
Sports game summarization aims at generating sports news from live commentaries.
However, existing datasets are all constructed through automated collection and cleaning processes, resulting in a lot of noise. Besides, current works neglect the knowledge gap between live commentaries and sports news, which limits the performance of sports game summarization.
In this paper, we introduce \textsc{K-SportsSum}, a new dataset with two characteristics:
(1) \textsc{K-SportsSum} collects a large amount of data from massive games. It has 7,854 commentary-news pairs. To improve the quality, \textsc{K-SportsSum} employs a manual cleaning process;
(2) Different from existing datasets, to narrow the knowledge gap, \textsc{K-SportsSum} further provides a large-scale knowledge corpus that contains the information of 523 sports teams and 14,724 sports players.
Additionally, we also introduce a knowledge-enhanced summarizer that utilizes both live commentaries and the knowledge to generate sports news.
Extensive experiments on \textsc{K-SportsSum} and \textsc{SportsSum} datasets show that our model achieves new state-of-the-art performances.
Qualitative analysis and human study further verify that our model generates more informative sports news.
\end{abstract}

\begin{CCSXML}
<ccs2012>
<concept>
<concept_id>10002951.10003317.10003347.10003357</concept_id>
<concept_desc>Information systems~Summarization</concept_desc>
<concept_significance>500</concept_significance>
</concept>
</ccs2012>
\end{CCSXML}

\ccsdesc[500]{Information systems~Summarization}

\keywords{datasets, sports game summarization, text summarization}


\maketitle

{%\par
  \medskip\small\noindent{\bfseries ACM Reference Format:}\par\nobreak
  \noindent\bgroup\def\\{\unskip{}, \ignorespaces}{Jiaan Wang, Zhixu Li, Tingyi Zhang, Duo Zheng, Jianfeng Qu, An Liu, Lei Zhao, and Zhigang Chen}\egroup. 2022. Knowledge Enhanced Sports Game Summarization. In \textit{Proceedings of the Fifteenth ACM International Conference on Web Search and Data Mining (WSDM '22), February 21--25, 2022, Tempe, AZ, USA}\textit{.} ACM, New York, NY, USA, \ref{TotPages}~pages. https://doi.org/10.1145/3488560.3498405
  }

\section{Introduction}
% Text Summarization aims at compressing the original document into a shorter text while preserving the main ideas.~\cite{rush-etal-2015-neural,chopra-etal-2016-abstractive,Nallapati2016AbstractiveTS,See2017GetTT,chen-bansal-2018-fast}. Recently, neural summarizers~\cite{Zhong2019SearchingFE,Liu2019TextSW,Xu2019DiscourseAwareNE,xu-durrett-2019-neural,lebanoff-etal-2019-scoring,cho-etal-2019-improving,Zhong2020ExtractiveSA,Wang2020HeterogeneousGN,jia-etal-2020-neural} have achieved impressive performance on general news domain datasets, such as CNN/DailyMail~\cite{Hermann2015TeachingMT,Nallapati2016AbstractiveTS} and XSum\cite{narayan-etal-2018-dont}. Nevertheless, it is nontrivial to transfer its success to other domains given that different domain texts have different characteristics.

\begin{figure}[t]
\setlength{\belowcaptionskip}{-12pt}
\centerline{\includegraphics[width=0.45\textwidth]{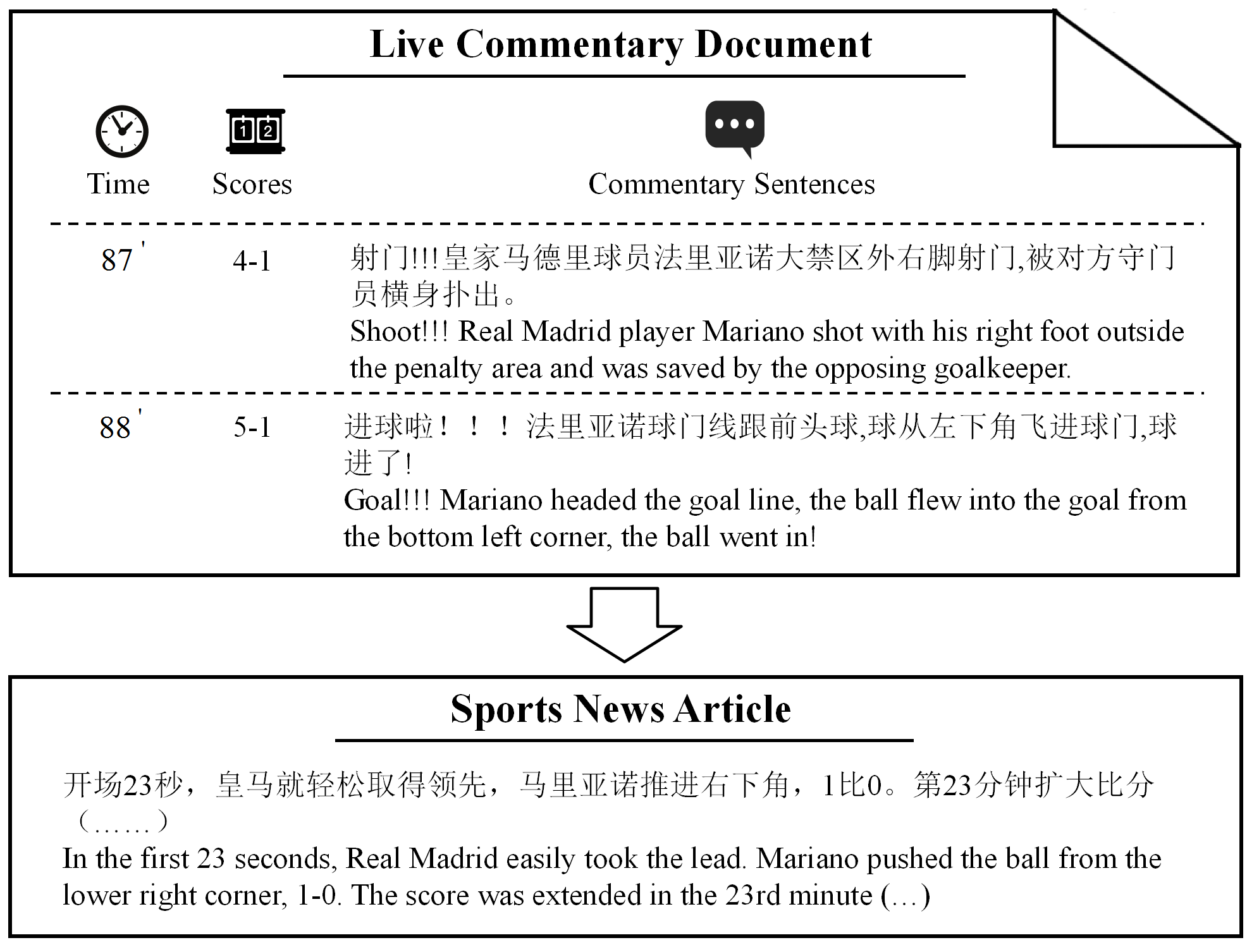}}
\caption{An example of Sports Game Summarization.}
\label{example}
\end{figure} 

In recent years, a large number of sports games are carried out every day, and it is demanding to report corresponding news articles after games. Meanwhile, manually writing sports news is labor-intensive for professional editors.
Therefore, how to automatically generate sports news has gradually attracted attention from both the research communities and industries.
As the example shown in Fig.~\ref{example}, Sports Game Summarization aims at generating sports news articles based on live commentaries~\cite{zhang-etal-2016-towards}. Ideally, the generated sports news should record the core events of a game that could help people efficiently catch up to games.
Compared to the most prior work on traditional text summarization, the challenges of sports game summarization lie in three aspects:
(1) The live commentaries record the whole events of a game, usually reaching thousands of tokens that is far beyond the typical 512 token limits of BERT-style pre-trained models;
(2) The live commentaries have a different text style from the sports news. Specifically, commentaries are more informal and colloquial;
% (2) There are different text styles between source live commentaries and target sports news. Specifically, commentaries are more informal and colloquial than news.
%
(3) There is a knowledge gap between commentaries and news. Sports news usually contains additional knowledge of sports teams or players, which cannot be obtained from corresponding commentaries (we will discuss more in Sec.~\ref{sec:knowledge_corpus}).

Most existing works and datasets on sports game summarization treat the problem as a single document summarization task.
% Most existing works typically follow the assumption that \textbf{all information in sports news can be inferred from the corresponding live commentaries}.
% Therefore, how to automatically generate sports news has gradually attracted attention from both the research communities and industries.
% Researchers explore various ways to generate sports news only based on the live commentaries.
%
Zhang et al.~\cite{zhang-etal-2016-towards} discuss this task for the first time and build the first dataset which has 150 commentary-news pairs. Wan et al.~\cite{Wan2016OverviewOT} also contribute a dataset with 900 samples for NLPCC 2016 shared task. Early methods design diverse strategies to select key commentary sentences, and then either form the sports news directly~\cite{zhang-etal-2016-towards,Zhu2016ResearchOS,Yao2017ContentSF} or relies on human-constructed templates to generate final news~\cite{Liu2016SportsNG,lv2020generate}.
Recently, Huang et al.~\cite{Huang2020GeneratingSN} present \textsc{SportsSum}, the first large-scale sports game summarization dataset with 5,428 samples. They crawl live commentaries and sports news from sports reports websites, and further adopt a rule-based cleaning process to clean the data.
Specifically, they first remove all HTML tags, and then for each news article, remove the descriptions before starting keywords which indicate the start of a game (e.g., ``at the beginning of the game'').
This is because there usually exist descriptions (e.g., matching history) at the beginning of news articles, which cannot be directly inferred from the corresponding commentaries.
% automated collection and rule-based cleaning processes
Huang et al.~\cite{Huang2020GeneratingSN} also extend early methods~\cite{zhang-etal-2016-towards,Zhu2016ResearchOS,Yao2017ContentSF,Liu2016SportsNG,lv2020generate} to a state-of-the-art two-step summarization framework, where each selected key commentary sentence is further rewritten to a news sentence through seq2seq models.

\begin{figure}[t]
\centerline{\includegraphics[width=0.48\textwidth]{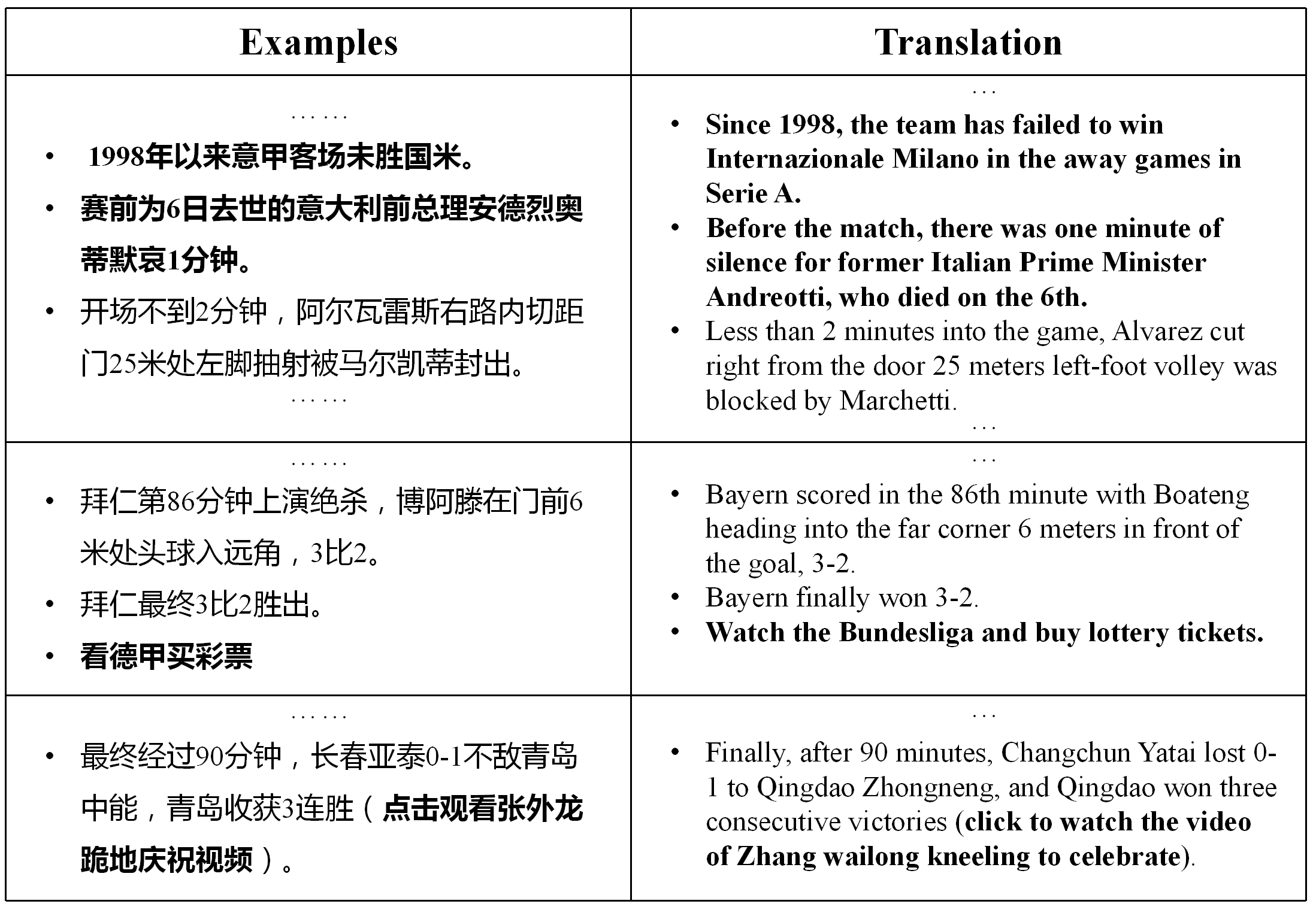}}
\setlength{\belowcaptionskip}{-10pt}
\caption{Noisy samples existed in \textsc{SportsSum} dataset. The first example contains the irrelevant descriptions to current game while the second one includes an advertisement. The last case has irrelevant hyperlink text.}
\label{noisy_samples}
\end{figure}

Despite above progress, there are two shortcomings in the previous works.
First, the previous datasets are limited in either scale or quality\footnote{We note that another concurrent work \textsc{SportsSum2.0}~\cite{Wang2021SportsSum20GH} also employs manual cleaning process on large-scale sports game summarization dataset. However, they only clean the original \textsc{SportsSum}~\cite{Huang2020GeneratingSN} dataset. There are three major differences between our dataset and \textsc{SportsSum2.0}: (1) the scale of our dataset is 1.45 times theirs; (2) our manual cleaning process also remove the history-related descriptions in the middle of news articles which is neglected by \textsc{SportsSum2.0}; (3) our dataset also provide a large-scale knowledge corpus to alleviate the knowledge gap issue.}. The scale of early datasets~\cite{zhang-etal-2016-towards,Wan2016OverviewOT} is less than 1,000 samples, which cannot be utilized to explore sophisticated supervised models.  \textsc{SportsSum}~\cite{Huang2020GeneratingSN} is many times larger than early datasets, but as shown in Fig.~\ref{noisy_samples}, we find more than 15\% of news articles in \textsc{SportsSum} have noisy sentences due to its simple rule-based cleaning process (detailed in Sec.~\ref{section:human_cleaning}).
Second, the knowledge gap between live commentaries and sports news is neglected by previous works. Though \textsc{SportsSum} removes a part of descriptions at the beginning of news to alleviate the knowledge gap to some extent, there are still many other descriptions leading to the knowledge gap. Moreover, their two-step framework neglects the gap.

In this paper, we introduce \textsc{K-SportsSum}, a large-scale human-cleaned sports game summarization dataset which is constructed with the following features:
(1) In order to improve both the scale and the quality of dataset, \textsc{K-SportsSum} collects a large amount of data from massive games and further employs a strict manual cleaning process to denoise news articles.
It is a large scale sports game summarization dataset, which consists of 7,854 high quality commentary-news pairs;
(2) To narrow the knowledge gap between live commentaries and sports news, \textsc{K-SportsSum} also provides an abundant knowledge corpus including the information of 523 sports teams and 14,724 sports players.
Additionally, in the aspect of model design, we propose a knowledge-enhanced summarizer that first selects key commentary sentences, and then considers the information of the knowledge corpus during rewriting each selected sentence to a news sentence so as to form final news.
% We evaluate our model on the \textsc{K-SportsSum} and \textsc{SportsSum}~\cite{Huang2020GeneratingSN} datasets.
%
The experimental results on \textsc{K-SportsSum} and \textsc{SportsSum}~\cite{Huang2020GeneratingSN} datasets show that our model achieves new state-of-the-art performances. We further conduct qualitative analysis and human study to verify that our model generates more informative sports news.

We highlight our contributions as follows:
\begin{itemize}[leftmargin=*,topsep=0pt]
\item We introduce a new sports game summarization dataset, i.e., \textsc{K-SportsSum}, which contains 7,854 human-cleaned samples. To the best of our knowledge, \textsc{K-SportsSum} is currently the highest quality and largest sports game summarization dataset$\footnote{We release the data at \url{https://github.com/krystalan/K-SportsSum}}$.
\item In order to narrow the knowledge gap between commentaries and news, we also provide an abundant knowledge corpus containing the information of 523 sports teams and 14,724 sports players.
%We hope this corpus could foster investigations into more sophisticated models for sports game summarization.
\item A knowledge-enhanced summarizer is proposed to take the information of knowledge corpus into account when generating sports news. It is the first sports game summarization model which considers the knowledge gap issue.
\item The experimental results show our model achieves a new state-of-the-art performance on both \textsc{K-SportsSum} and \textsc{SportsSum} datasets. Qualitative analysis and human study further verify that our model generates better sports news.
\end{itemize}

\section{Data Construction}
In this section, we first show how we collect live commentary documents and news articles from Sina Sports Live (\S~\ref{sec:data_collection}).
Secondly, we analyze the noise existed in collected text and introduce the manual cleaning process to denoise the news articles (\S~\ref{section:human_cleaning}).
Thirdly, we discuss the collection process of knowledge corpus (\S~\ref{sec:knowledge_corpus}).
Finally, we give the details of benchmark settings (\S~\ref{sec:benchmark_settings}).

\subsection{Data Collection}
\label{sec:data_collection}
% Previous datasets~\cite{zhang-etal-2016-towards,Wan2016OverviewOT,Huang2020GeneratingSN} are all in Chinese since the live text commentary services are extremely popular in China, where sports fans in many cases do not have access to live video streams due to copyright restriction.
Following previous works~\cite{zhang-etal-2016-towards,Wan2016OverviewOT,Huang2020GeneratingSN}, we crawl the records of football games from Sina Sports Live\footnote{\url{http://match.sports.sina.com.cn/index.html}}, the most influential football live services in China.
Note that the existing research of sports game summarization is oriented to the football games which are the easiest to collect, but the methods and discussions can trivially generalize to other types of sports games.
After crawling all football games data from 2012 to 2020, we remove HTML tags and obtain 8,640 live commentary documents together with corresponding news articles.

\begin{table*}[t]
  \centering
  \setlength{\belowcaptionskip}{-10pt}
  \resizebox{0.80\textwidth}{!}
  {
    \begin{tabular}{c|c|l}
      \hline
      \textbf{Additional Knowledge}                     & \%                    & \multicolumn{1}{c}{\textbf{Examples}}                                                                            \\ \hline
      \multirow{2}{*}{\textbf{Home or visiting team}} & \multirow{2}{*}{14.7} & \textbf{The home team} broke the deadlock in the 58th minute, Brian took a free kick and Lucchini scored     \\
                                                   &                       & the goal. In the 77th minute, Nica...                                                                       \\ \hline
      \multirow{4}{*}{\textbf{Player Information}}          & \multirow{4}{*}{6.3}  & Paris exceeded it again in the 26th minute! Motta took a corner kick from the right. Verratti, \textbf{who is}   \\
                                                   &                       & \textbf{1.65 meters tall}, pushed into the near corner at a small angle 6 meters away from the goal, 2-1. ...    \\ \cline{3-3}
                                                %   &                       & In the 54th minute, Darmstadt defender Sirigu made a tackle in the penalty area. Tomorrow, Pulisic, \\
                                                %   &                       & \textbf{who will celebrate his 18th birthday}, pushes into the net from the right side, 3-0! ...                     \\ \cline{3-3} 
                                                   &                       & After another 3 minutes, Keita broke through Campagnaro and Brugman and passed back. Immobile   \\
                                                   &                       & shot to the top of the goal. He faced the \textbf{previous teams} without celebrating the goal. ...                      \\ \hline
      \multirow{3}{*}{\textbf{Team Information}}            & \multirow{3}{*}{4.7}  & \textbf{Purple Lily} equalized in the 29th minute, Tomovic broke through Ljajic and Duoduo from the right.   \\ \cline{3-3} 
                                                   &                       & \textbf{The last Champions League runner-up} completely controlled the game after the opening. Reus   \\
                                                   &                       & shot higher once with two feet and was rescued by Mandanda once. ...                                                \\ \hline
      \end{tabular}
  }
  \setlength{\belowcaptionskip}{3pt}
  \caption{Additional knowledge required for sports game summarization.}
  \label{table:addition_knowledge}
\end{table*}

\subsection{Data Cleaning}
\label{section:human_cleaning}

\subsubsection{Noise Analysis}
We find that the \emph{live commentaries} are of high-quality due to the \emph{structured form}. Nevertheless, the \emph{news articles} are \emph{unstructured text} usually containing noises. Specifically, we divide all noises into four types:

\begin{itemize}[leftmargin=*,topsep=0pt]
\item Description of other games: We find about 12\% of the news pages contain multiple news articles belonging to different games, which has been neglected by \textsc{SportsSum}~\cite{Huang2020GeneratingSN}, resulting in 2.2\% (119/5428) of news articles include descriptions of other games.

\item Advertisements: There are advertisements often appear in news articles. We find about 9.3\% (505/5428) of news articles in \textsc{SportsS\\um} have such noise.

\item Irrelevant hyperlink texts: Many new sentences contain hyperlink texts. Some of them can be regarded as part of the news, but others are irrelevant to current news, which we called irrelevant hyperlink texts. About 0.6\% (31/5428) samples of \textsc{SportsSum} have this kind of noise.

\item History-related descriptions: There are amount of news articles containing history-related descriptions which cannot be inferred from the corresponding live commentary document.
\textsc{SportsSum} adopts a rule-based approach to remove all news sentences before starting keywords to alleviate part of this noise.
However, this approach cannot correctly dispose about 4.6\% (252/5428) of the news articles, because the rule-based approach cannot cover all situations.
In addition, history-related descriptions may not only appear at the beginning of the news. Many news articles often introduce the matching history in the middle, e.g., if a player scores, the news may introduce the recent outstanding performance of the player in the previous rounds, or count his (her) total goals in the current season.
% Besides, some news articles may summarize the impact of the competition results on both sides or forecast future games in the end. This information also cannot be inferred from the corresponding commentaries, but has been ignored by previous work.
%
We also explain why we consider the history-related descriptions as noise rather than the knowledge gap between commentaries and news in Sec.~\ref{sec:knowledge_corpus}.
\end{itemize}

\begin{figure}[t]
\centering
\setlength{\belowcaptionskip}{-10pt}
\centerline{\includegraphics[width=0.45\textwidth]{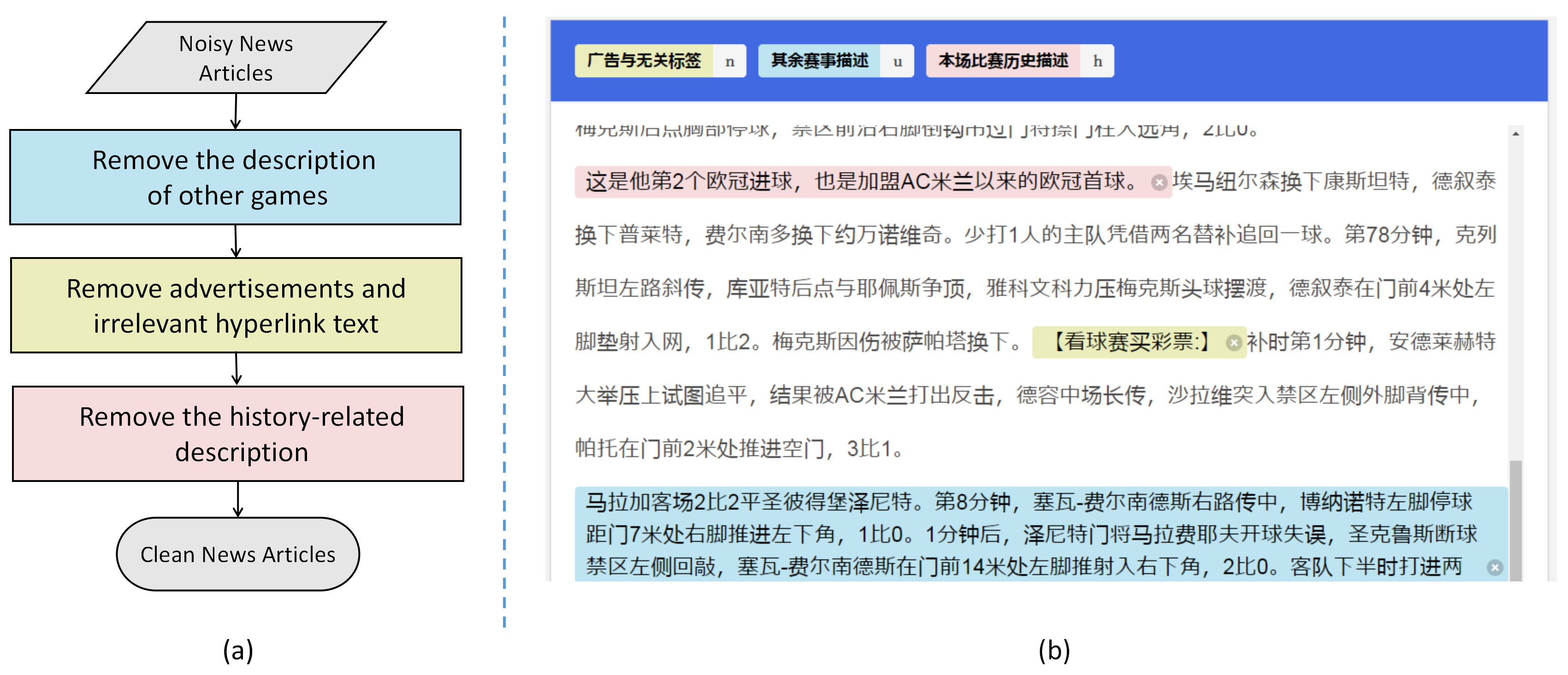}}
\caption{(a) Flow chart of manual cleaning process. (b) Screenshot of manual annotation interface.}
\label{fig:human}
\end{figure}

\subsubsection{Manual Cleaning Process}
To reduce the noise of news articles, we design a manual cleaning process with special consideration for sports game summarization.
Fig.~\ref{fig:human}(a) shows the manual cleaning process. We first remove the description of other games. Secondly, we delete advertisements and irrelevant hyperlink text. Finally, the history-related descriptions of the game are identified and removed.

\subsubsection{Annotation Process}
The interface of manual cleaning is shown in Fig.~\ref{fig:human}(b).
We recruit 9 master students, who are native Chinese speakers to perform the manual cleaning process. Firstly, we randomly select 50 news articles as test samples and ask all students to clean them at the same time. Based on the results, we decide 7 annotators and 2 senior annotators. After that, each news article will be randomly assigned to 2 annotators. If the cleaning results are inconsistent, it will be determined by a third senior annotator. Finally, all the cleaning results are checked by another two data experts. If the data experts think that the result does not meet the requirements, the news article will be assigned again.

\subsubsection{Post-processing}
After the manual cleaning process, we discover some news articles do not contain the description of the current game, or contain little information, e.g., only include the results of the game.
We discard these news articles and retain 7,854 high-quality manually cleaned news articles which together with the corresponding live commentary documents constitute the \textsc{K-SportsSum} dataset.

\subsection{Knowledge Corpus}
\label{sec:knowledge_corpus}
As we mentioned in Sec.~\ref{section:human_cleaning}, the history-related descriptions in original news articles are regarded as noise and have been removed during the manual cleaning process due to the following reasons: (1) we follow the settings of \textsc{SportsSum}~\cite{Huang2020GeneratingSN} which adopt a rule-based approach to remove all sentences before the starting keywords. Most of the removed sentences are history-related descriptions; (2) the goal of sports game summarization is to generate news articles that can \emph{record the key events of the \textbf{current} sports games}. The history-related descriptions make little contribution to the goal.

To investigate if there are still other descriptions in the news articles, which also cannot be inferred from the corresponding live commentary document, we randomly select 300 samples from \textsc{K-SportsSum} and manually analyze whether the news articles contain additional knowledge that cannot be obtained from commentary documents. Tab.~\ref{table:addition_knowledge} shows statistics of required additional knowledge: (1) Some news articles replace the name of football teams with ``home team'' or ``visiting team'', which cannot be explicitly obtain from commentary documents; (2) Some news includes the personal information of players, e.g., height, birthday, previous teams; (3) A number of news articles contain prior knowledge of football teams, such as nickname (``Purple Lily'' is the nickname of ACF Fiorentina, which is shown in the penultimate line of Tab.~\ref{table:addition_knowledge}) and past honors.

To narrow the knowledge gap between commentaries and news, we construct a knowledge corpus whose collection process contains the following three steps:

\begin{figure}[t]
\centering
\setlength{\belowcaptionskip}{-10pt}
\centerline{\includegraphics[width=0.45\textwidth]{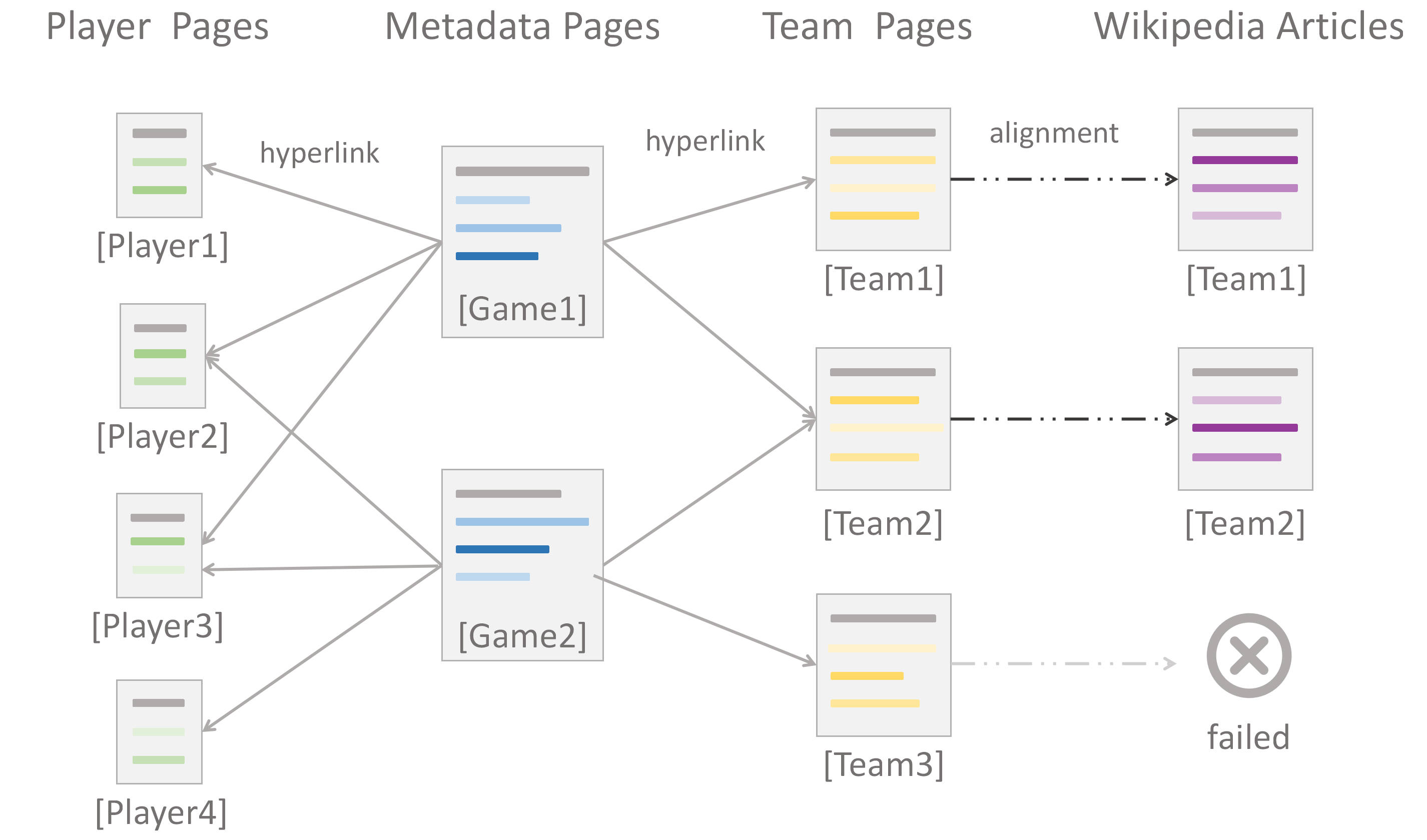}}
\caption{The relations among metadata pages, team pages, player pages and Wikipedia articles. Each game has a metadata page that can link to related player or team pages. We further align each team page to the corresponding Wikipedia article.}
\label{fig:pages}
\end{figure}

\vspace{1ex}
\noindent\textbf{Step 1: Metadata Collection.}
For each game, as shown in Fig.~\ref{fig:pages}, Sina Sports Live also provides a \emph{metadata page} that can link to related \emph{player pages} and \emph{team pages}. Note that each team (or player) has a unique corresponding team (or player) page.
After crawling 8,640 metadata pages of all games in \textsc{K-SportsSum}, we obtain 559 URLs of team pages and 15591 URLs of player pages.

\vspace{1ex}
\noindent\textbf{Step 2: Player Knowledge Collection.}
The player pages provided by Sina Sports Live contain structured knowledge cards describing players in ten aspects (i.e., name, birthday, age, etc.).
We crawl these pages and obtain the 14,724 players' structured knowledge cards.
%\footnote{A small number of player pages are invalid, so the number of collected knowledge card is less than the number of player pages.}
Then we convert each knowledge card to a passage through several rule-based sentence templates (e.g., an item of knowledge card $\langle$``Ronaldo'', ``birthday'', ``September 18, 1976''$\rangle$ can be converted to a sentence ``Ronaldo's birthday is September 18, 1976''). In this way, we obtain 14,724 player passages.

\vspace{1ex}
\noindent\textbf{Step 3: Team Knowledge Collection.}
Though Sina Sports Live offers the team pages, we find that most of them are less informative than player pages. To construct an informative knowledge corpus, we decide to manually align these 559 team pages to Wikipedia articles\footnote{https://zh.wikipedia.org/} in which we further extract plain text\footnote{We crawl the Wikipedia articles and then use wikiextractor tool (\url{https://github.com/attardi/wikiextractor}) to extract plain text.} to form our knowledge corpus.
Three master students and two data experts are recruited to perform the alignment task. Each team page is assigned to three students, if their results are inconsistent, the final result is decided by a group meeting with these five persons. Eventually, we assign 523 out of 559 team pages to corresponding Wikipedia articles.

Finally, our knowledge corpus contains 523 team articles and 14,724 player passages. The corpus also provides the link and alignment relations among metadata pages, player pages, team pages and Wikipedia articles. Thus, for a given game, we can accurately retrieve the related articles and passages.

\begin{figure}[t]
\setlength{\belowcaptionskip}{-10pt}
  \centerline{\includegraphics[width=0.45\textwidth]{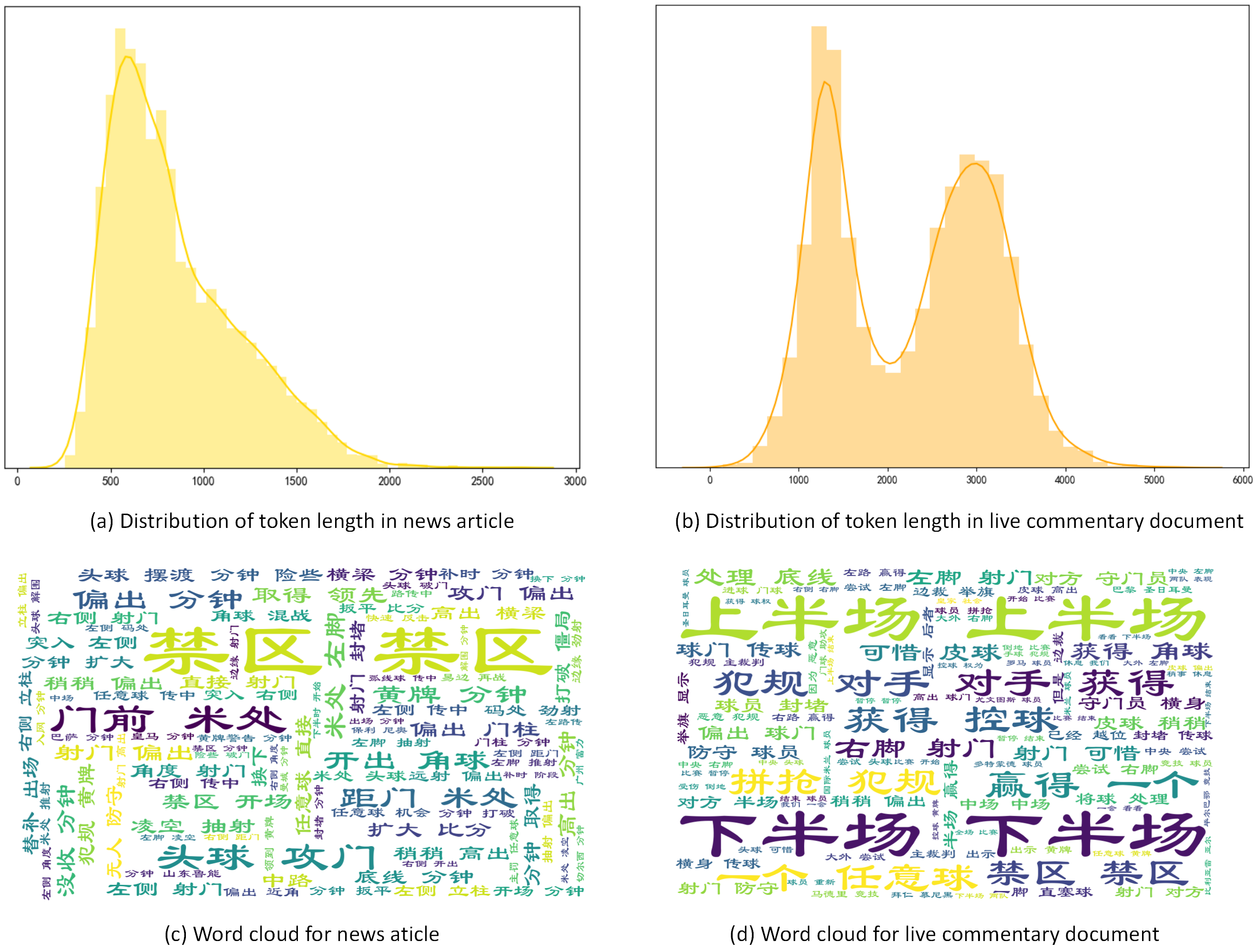}}
  \caption{Statistics of tokens in news article and commentary document. Stop words were excluded from word clouds.}
  \label{fig:deep_statistics}
\end{figure}

\begin{table*}[t]
  \centering
  \setlength{\belowcaptionskip}{-10pt}
  \resizebox{0.95\textwidth}{!}
  {
    \begin{tabular}{l|l|cccccc|cccccc}
      \hline
      \multicolumn{1}{c|}{\multirow{3}{*}{Datasets}} & \multicolumn{1}{c|}{\multirow{3}{*}{\# Examples}} & \multicolumn{6}{c|}{News article}                                                             & \multicolumn{6}{c}{Live commentary document}                                                \\
      \multicolumn{1}{c|}{}                         & \multicolumn{1}{c|}{}                             & \multicolumn{2}{c}{\# Tokens} & \multicolumn{2}{c}{\# Words} & \multicolumn{2}{c|}{\# Sent.} & \multicolumn{2}{c}{\# Tokens} & \multicolumn{2}{c}{\# Words} & \multicolumn{2}{c}{\# Sent.} \\
      \multicolumn{1}{c|}{}                         & \multicolumn{1}{c|}{}                             & Avg.         & 95th pctl.      & Avg.        & 95th pctl.     & Avg.        & 95th pctl.      & Avg.         & 95th pctl.      & Avg.        & 95th pctl.     & Avg.        & 95th pctl.     \\ \hline
      Zhang et al.~\cite{zhang-etal-2016-towards}                                  & 150                                               & -            & -               & -           & -              & -           & -               & -            & -               & -           & -              & -           & -              \\ \hline
      NLPCC 2016 shared task~\cite{Wan2016OverviewOT}                        & 900                                               & -            & -               & -           & -              & -           & -               & -            & -               & -           & -              & -           & -              \\ \hline
      \textsc{SportsSum}\cite{Huang2020GeneratingSN}                                     & 5428                                              & 801.11       & 1558               & 427.98      & 924              & 23.80        & 39               & 3459.97      & 5354               & 1825.63     & 3133              & 193.77      & 379              \\ \hline
      \textsc{K-SportsSum}                        & \textbf{7854}                                              & 606.80        & 1430            & 351.30       & 845            & 19.22       & 40              & 2251.62      & 3581            & 1200.31     & 1915           & 187.69      & 388            \\
      \ \ \ \textsc{K-SportsSum} (train)             & 6854                                              & 604.41       & 1427             & 349.88       & 843            & 19.31       & 40              & 2250.04      & 3581            & 1199.69     & 1915           & 187.81      & 390            \\
      \ \ \ \textsc{K-SportsSum} (dev.)                & 500                                               & 615.04      & 1426            & 356.51      & 824            & 19.64       & 42              & 2280.04      & 3608            & 1216.76      & 1920           & 187.08      & 370            \\
      \ \ \ \textsc{K-SportsSum} (test)               & 500                                               & 631.28      & 1463            & 365.54      & 859            & 20.10       & 41              & 2244.78      & 3563            & 1192.32      & 1917           & 186.73      & 376            \\ \hline
      \end{tabular}
  }
  \setlength{\belowcaptionskip}{3pt}
  \caption{Statistics of \textsc{K-SportsSum} and previous datasets (Sent.: sentence, Avg.: average, 95th pctl.: 95th percentile).}
  \label{table:statistic}
\end{table*}

\begin{figure*}[t]
\centerline{\includegraphics[width=0.85\textwidth]{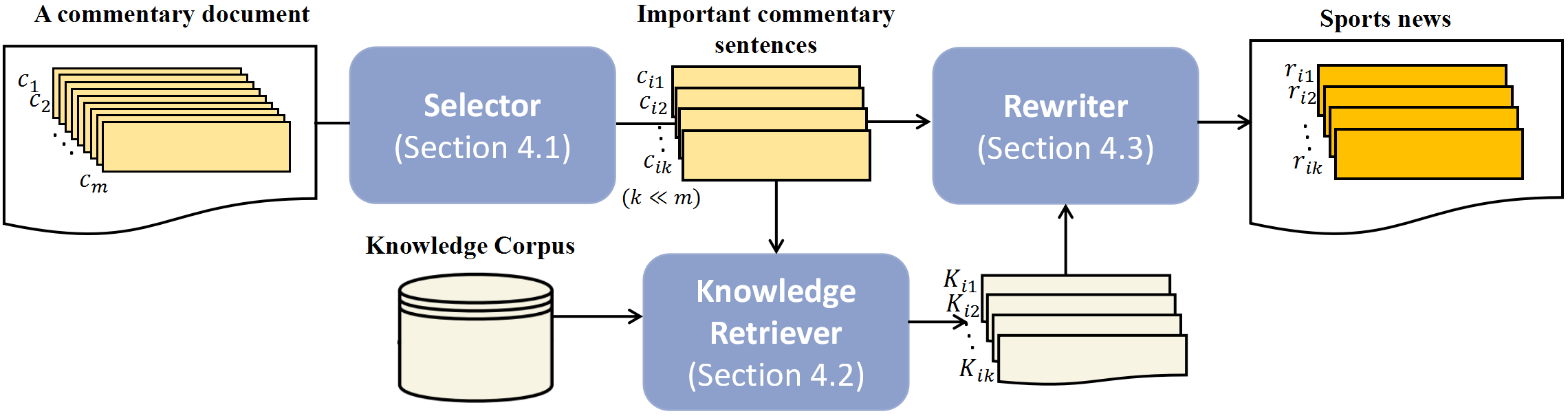}}
\caption{An overview of knowledge-enhanced summarizer.}
\label{fig:overview}
\end{figure*}

\subsection{Benchmark Settings}
\label{sec:benchmark_settings}
\vspace{0.5ex}
We randomly select 500 samples and other 500 samples from \textsc{K-SportsSum} to form development set and testing set.
The remaining 6,854 samples constitute training set.

\section{Data Analysis}
In this section, we analyze various aspects of \textsc{K-SportsSum} to provide a deeper understanding of the dataset and task of sports game summarization.

\vspace{1ex}
\noindent\textbf{Data Size.} Tab.~\ref{table:statistic} shows the statistics of \textsc{K-SportsSum} and previous datasets. The \textsc{K-SportsSum} dataset is much larger than any other dataset.

\vspace{1ex}
\noindent\textbf{News Article.} The average number of tokens in \textsc{K-SportsSum} is 606.80 which is less than the counterpart in \textsc{SportsSum} (801.11) due to the manual cleaning process.
Fig.~\ref{fig:deep_statistics}(a) shows the distributions of token length for news articles.
This distribution obeys positive skewness distribution which indicates the length of most news articles is less than the average.

\vspace{1ex}
\noindent\textbf{Live Commentary Document.} The length of commentary document usually reaches thousands of tokens, which makes the task more challenging.
The average number of tokens for commentary documents in \textsc{SportsSum} is 3459.97, which is longer than the counterpart in \textsc{K-SportsSum}, i.e., 2251.62.
This is because \textsc{SportsSum} considers every commentary sentences in the document, but we only retain the commentary sentences which has timeline information, for the reason that most the commentary sentences without timeline information are irrelevant to the current game.
The distributions of token length for live commentary documents in \textsc{K-SportsSum} are shown in Fig.~\ref{fig:deep_statistics}(b). This distribution obeys the Gaussian mixture distribution because the live services in Sina Sports Live have been updated once. The length distribution of commentary documents is different before and after the update. Some commentary documents in \textsc{K-SportsSum} are provided by the old live services, while others are offered by the new one.

\vspace{1ex}
\noindent\textbf{Word Clouds.}
Fig.~\ref{fig:deep_statistics}(c) and Fig.~\ref{fig:deep_statistics}(d) show the word clouds for news article and live commentary document, where is easy to find the different text styles between these two types of text.

\vspace{1ex}
\noindent\textbf{Knowledge Corpus.}
The knowledge corpus contains 523 team articles and 14,724 player passages.
Each team article has 18.28 sentences or 1341.91 tokens on average. Each player passage contains 15.05 sentences or 283.49 tokens on average.

\section{Model}
We propose knowledge-enhanced summarizer which generates sports news based on both live commentary document and knowledge corpus. Formally, a model is given a live commentary document $C=\{(t_{1},s_{1},c_{1}),...,(t_{m},s_{m},c_{m})\}$ together with a knowledge corpus $K$ and outputs a sports news article $R=\{r_{1},r_{2},...,r_{n}\}$. $r_{i}$ represents $i$-th news sentence and $(t_{j},s_{j},c_{j})$ is $j$-th commentary, where $t_{j}$ is the timeline information, $s_{j}$ denotes the current scores and $c_{j}$ is the commentary sentence.
% As shown in Fig.~\ref{example}, the goal of sports game summarization is to generate sports news $R=\{r_{1},r_{2},...,r_{n}\} $ from a given live commentary document $C=\{(t_{1},s_{1},c_{1}),...,(t_{m},s_{m},c_{m})\}$ $(m \geq n)$.  $r_{i}$ represents $i$-th news sentence and $(t_{j},s_{j},c_{j})$ is $j$-th commentary, where $t_{j}$ is the timeline information, $s_{j}$ denotes the current scores and $c_{j}$ is the commentary sentence.

The overview of our knowledge-enhanced summarizer is illustrated in Fig.~\ref{fig:overview}. Firstly, we utilize a selector to extract key commentary sentences from the original commentary document (Sec.~\ref{sec:selector}). Secondly, a knowledge retriever is used to obtain related passages and articles from the knowledge corpus for each selected commentary sentence (Sec.~\ref{sec:retriever}). Lastly, we make use of a seq2seq rewriter to generate news sentences based on the corresponding commentary sentences and retrieved passages/articles (Sec.~\ref{sec:rewriter}). In order to train our selector and rewriter, we need labels to indicate the importance of commentary sentences and aligned $\langle$commentary sentence, news sentence$\rangle$ pairs.
Following Huang et al.~\cite{Huang2020GeneratingSN}, we obtain these importance labels and sentence pairs through a sentence mapping process (Sec.~\ref{sec:sentence_pairs}).

\subsection{Key Commentary Sentence Selection}
\label{sec:selector}
The selector is used to extract key commentary sentences $\{c_{i1},c_{i2},...\\,c_{ik}\}$ from all commentary sentences $\{c_{1},c_{2},...,c_{m}\}$ of a given live commentary document, which can be modeled as a text classification task.
Different from the previous state-of-the-art two-step framework~\cite{Huang2020GeneratingSN} purely utilizing TextCNN~\cite{Kim2014ConvolutionalNN} as the selector and ignores the contexts of a commentary sentence, we make use of RoBERTa~\cite{Liu2019RoBERTaAR} to extract the contextual representation of a commentary sentence, and then predict its importance.
Specifically, we input the target commentary sentence with its context to RoBERTa~\cite{Liu2019RoBERTaAR} in a sliding window way. The representation of target sentence is obtained by averaging the output embedding of each token belonging to the target sentence.
Finally, a sigmoid classifier is employed to predict the importance of the target commentary sentence.
The cross-entropy loss is used as the training objective for the selector.

\begin{figure*}[t]
\centerline{\includegraphics[width=0.80\textwidth]{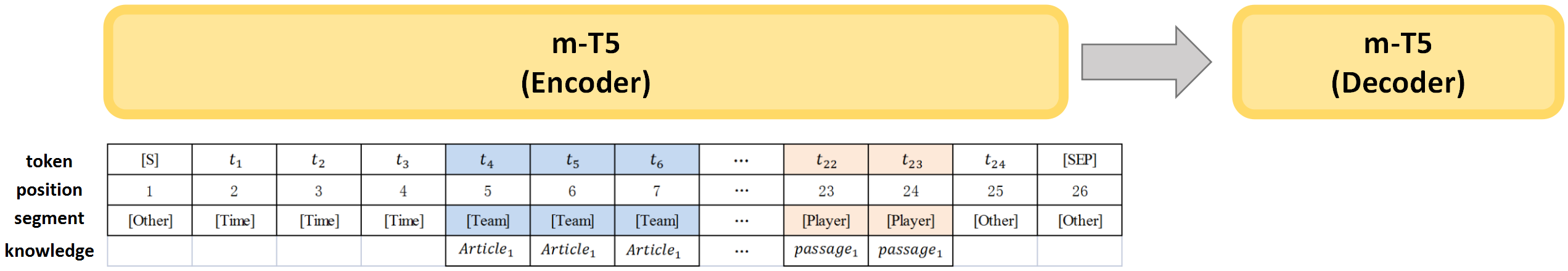}}
\caption{Our encoder-decoder rewriter architecture.}
\label{fig:rewriter}
\end{figure*}

\subsection{Knowledge Retrieval}
\label{sec:retriever}
Given a selected commentary sentence $c_{ij}$ and the knowledge corpus $K$, the knowledge retriever first recognize the team and player entity mentions in $c_{ij}$, and then link the entity mentions to team articles or player passages from the corpus $K$.

\subsubsection{Named Entity Recognition Process}
In order to recognize the players and teams mentioned in commentary sentence $c_{ij}$, we train a FLAT model~\cite{Li2020FLATCN} (a state-of-the-art Chinese NER model which could better leverage the lattice information of Chinese characters sequences) on MSRA~\cite{Levow2006TheTI} (a general Chinese NER dataset), and then make use of the trained FLAT model to predict the entity mentions in $c_{ij}$. We only retain the \texttt{PER} and \texttt{ORG} entity mentions predicted by FLAT.
Because the \texttt{PER} entity mentions indicate the players while the \texttt{ORG} entity mentions hint the teams.
%which we denote as $\{p_{1},p_{2},...,p_{s}\}$ and $\{o_{1},o_{2},...,o_{t}\}$, respectively.

\subsubsection{Entity Linking Process}
For a given game, we can accurately retrieve dozens of candidate player passages and several (2 in most cases) candidate team articles through the link and alignment relations provided by knowledge corpus.
%that describe the players and teams who participate in the current game. Thus we can obtain dozens of (less than 40 in most instances) candidate player passages and several (2 in most cases) candidate team articles. 
The entity linking process needs to link each \texttt{PER} or \texttt{ORG} entity mention to the candidate passages/articles. Here, we employ a simple yet effective linking method, we calculate the normalized Levenshtein distance between each entity mention with the title\footnote{The standard name of each player is regarded as the title of player passages, while the team articles collected from Wikipedia already has the corresponding titles.} of passages/articles:

\begin{equation}
\label{levenshtein}
NLev(entity,title)=\frac{Lev(entity,title)}{max(len(entity),len(title))}
\end{equation}
where $NLev(\cdot,\cdot)$ means the normalized Levenshtein distance, and $Lev(\cdot,\cdot)$ represents the standard Levenshtein distance. $len(\cdot)$ indicates the number of characters within input sequences. 

Each \texttt{PER} (or \texttt{ORG}) entity mention is linked to the corresponding player passage (or team article) whose title has the nearest normalized Levenshtein distance with the entity mention if the nearest distance within a predefined threshold $\lambda_{p}$ (or $\lambda_{o}$). Otherwise, we do not link the entity mention.
Eventually, for a given commentary sentence $c_{ij}$, we obtain a number of $\langle$linked \texttt{PER}/\texttt{ORG} entity mention, corresponding passage/article$\rangle$ pairs.
%which we denote as $\{p_{1},p_{2},...,p_{s}\}$ and $\{o_{1},o_{2},...,o_{t}\}$, respectively.
% Eventually, for a given commentary sentence $c_{ij}$, we obtain a linked entity mention set $E=\{p_{1},p_{2},...,p_{s}\} \cup \{o_{1},o_{2},...,o_{t}\}$.

\subsection{Sports News Generation}
\label{sec:rewriter}
As shown in Fig.~\ref{fig:rewriter}, we make use of mT5~\cite{Xue2021mT5AM}, a pre-trained multilingual language model\footnote{Since there is no Chinese version of T5 for public use, we choose its multilingual version, i.e., mT5.}, as our rewriter to generate a news sentence $r_{ij}$ based on a given commentary sentence $c_{ij}$, its timeline information $t_{ij}$ and the $\langle$linked \texttt{PER}/\texttt{ORG} entity mention, corresponding passage/article$\rangle$ pairs.
Specifically, we tokenize the temporal phrase ``In the $t_{ij}$-minute'' and $c_{ij}$ using mT5's tokenizer, then form the input as \texttt{<s> temporal phrase </s> commentary sentence </s>}. The input embeddings of each token consist of a segment embedding $z^{seg}$ and a knowledge embedding $z^{know}$ in addition to a token embedding $z^{token}$ and a position embedding $z^{pos}$ within the input sequence:
\begin{equation}
\label{equa:input_embedding}
z_{k} = LN(z_{k}^{token}+z_{k}^{pos}+z_{k}^{seg}+z_{k}^{know})
\end{equation}
where $z_{k}$ denotes the fused embedding of the $k$-th token in the input sequence. $LN(\cdot)$ represents layer normalization.

We explain the two additional embeddings below.

\noindent\textbf{Segment embeddings.} To convey the semantics of fine-grained types of tokens, we use four learnable segment embeddings (\texttt{[Playe\\r], [Team], [Time] and [Other]}) to indicate a token belongs to: (1) linked \texttt{PER} entity mentions; (2) linked \texttt{ORG} entity mentions; (3) the temporal phrase; (4) none of above.

\noindent\textbf{Knowledge embeddings.} For a token belonging to the linked \texttt{PER}/\texttt{ORG} entity mentions, we consider the representation of corresponding passage/article as its knowledge embedding. In detail, we first input each sentence of a given passage/article to a pre-trained RoBERTa~\cite{Liu2019RoBERTaAR} one by one and use the output embedding of \texttt{[CLS]} token as the sentence embedding. Then, the representation of the whole passage/article is the average of all sentences' embeddings.

The input embeddings are further passed to the mT5-encoder and generate the news sentence $r_{ij}$ in the sequence-to-sequence learning process with the negative log-likelihood loss.
All generated news sentences are concatenated to form final sports news.

\subsection{Training with Oracle}
\label{sec:sentence_pairs}
To train the selector and rewriter, we need (1) labels to indicate the importance of each commentary sentence and (2) a large number of $\langle$commentary sentence, news sentence$\rangle$ pairs, respectively.
To obtain the above labels and pairs, we \textbf{map each news sentence to its commentary sentence} through the same sentence mapping process as Huang et al.~\cite{Huang2020GeneratingSN}.
Specifically, although there is no explicit timeline information on news articles, we find about 40\% news sentences in \textsc{K-SportsSum} begin with ``in the n-th minutes''.
For each news sentence $r_{i}$ beginning with ``in the $h_{i}$-th minutes'', we first obtain its time information $h_{i}$. Then, we consider commentary sentences $c_{j}$ whose corresponding timeline $t_{j} \in [h_{i},h_{i}+3]$ as candidate mapping set of $r_{i}$. Lastly, we calculate BERTScore~\cite{Zhang2020BERTScoreET} (a metric to measure the sentence similarity) of $r_{i}$ and all the commentary sentences in candidate mapping set. The commentary sentence with the highest score is paired with $r_{i}$.
With the above process, we can obtain a large number of pairs of a mapped commentary sentence and a news sentence, which can be used to train our rewriter. Furthermore, the commentary sentence appearing in the pairs will be regarded as important while others are insignificant, which could be used as the training data for our selector.

\begin{table*}[t]
  \centering
  \resizebox{0.80\textwidth}{!}
  {
    \centering
    \begin{tabular}{clcccccc}
      \cline{1-8}
      \multirow{2}{*}{Method}                                                        & \multirow{2}{*}{Model} & \multicolumn{3}{c}{\textsc{K-SportsSum}}                                   & \multicolumn{3}{c}{\textsc{SportsSum}}                                      \\
                                                                                     &                        & ROUGE-1              & ROUGE-2              & ROUGE-L              & ROUGE-1              & ROUGE-2              & ROUGE-L              \\ \cline{1-8}
      \multirow{3}{*}{Extractive Models}                                             & TextRank               & 21.04                & 6.17                 & 20.64                & 18.37                & 5.69                 & 17.23                \\
      & PacSum                 & 24.33                & 7.28                 & 24.04                & 21.84                & 6.56                 & 20.19  \\
                                                                                     & Roberta-Large                 & 27.79                & 8.14                 & 27.21                & 26.64                & 7.44                 & 25.59                \\ \cline{1-8}
      \multirow{2}{*}{Abstractive Models}                                            & Abs-LSTM                   & 31.24                & 11.27                & 30.95                & 29.22                & 10.94                & 28.09                \\
                                                                                     & Abs-PGNet                  & 36.49                & 13.34                & 36.11                & 33.21                & 11.76                & 32.37                \\ \cline{1-8}
      \multirow{2}{*}{\makecell[c]{Two Step Framework \\ (Selector + Rewriter)}}                      & SportsSUM~\cite{Huang2020GeneratingSN}              & 44.89                & 19.04                & 44.16                & 43.17                & 18.66                & 42.27                \\
                                                                                     & KES (Our)                  & \textbf{48.79}                & \textbf{21.04}                & \textbf{47.17}                & \textbf{47.43}                & \textbf{20.54}                & \textbf{47.79}                \\ \cline{1-8}
                                      
      \end{tabular}
  }
  \caption{Experimental results on \textsc{K-SportsSum} and \textsc{SportsSum}.}
  \label{table:result}
\end{table*}

\begin{table*}[t]
  \centering
    \begin{tabular}{p{0.25\textwidth}p{0.03\textwidth}p{0.30\textwidth}p{0.32\textwidth}}
    \hline
    \multicolumn{1}{c}{Commentary Sentence} & \multicolumn{1}{c}{T.} & \multicolumn{1}{c}{KES (w/o know.)} & \multicolumn{1}{c}{KES} \\ \hline
    Barcelona attacked on the left side, Semedo passed the ball to the front of the small restricted area, Suarez scored the ball!!! & \multicolumn{1}{c}{27'}      & Barcelona broke the deadlock in the 27th minute, Semedo crossed from the left and Suarez scored in front of the small restricted area. & \textbf{Defending champion} broke the deadlock in the 27th minute, Semedo crossed from the left and Suarez scored in front of the small restricted area. \\ \hline
    De Yang was replaced and Song Boxuan came on as a substitute                                                                     & \multicolumn{1}{c}{45'}      & In the 45th minute, De Yang was replaced by Song Boxuan                                                                                & In the 45th minute, De Yang was replaced by Song Boxuan, \textbf{who is 1.7 meters height.}                                                               \\ \hline
    \end{tabular}
%   }
  \caption{Qualitative analysis on \textsc{K-SportsSum} development set (T.:timeline information)}
  \label{table:case_study}
\end{table*}

\section{Experiments}

\subsection{Implementation Details}
We train all models on one 32GB Tesla V100 GPU. Our knowledge-enhanced summarizer is implemented based on RoBERTa~\cite{Liu2019RoBERTaAR} and mT5~\cite{Xue2021mT5AM} of huggingface Transformers library~\cite{wolf-etal-2020-transformers} with default settings.
In detail, we utilize RoBERTa-Large (12 layers with 1024 hidden size) to initialize our selector, and mT5-Large (24 layers with 1024 hidden size) as our rewriter. Another fixed RoBERTa-Large in our rewriter is used to calculate the knowledge embedding for a number of input tokens.
For knowledge retriever, we train a FLAT~\cite{Li2020FLATCN} NER model on a general Chinese NER dataset, i.e., MSRA~\cite{Levow2006TheTI}. The trained FLAT model achieves 94.21 F1 score on the MSRA testing set, which is similar to the original paper. The predefined threshold $\lambda_{p}$ and $\lambda_{o}$ used in the entity linking process are 0.2 and 0.25, respectively.
We set the hyperparameters based on the preliminary experiments on the development set. We use minimal hyperparameter tuning using Learning Rates (LRs) in [1e-5, 2e-5, 3e-5] and epochs of 3 to 10. We find the selector with LR of 3e-5 and 5 epochs to work best. The best configuration for the mT5 rewriter is 2e-5 lr, 7epochs. For all models, we set the batch size to 32, use Adam optimizer with a default initial momentum and adopt linear warmup in the first 500 steps.
% For pointer-generator network rewriter, hyperparameter searches were minimal and consisted of grid searches of LR in [0.05, 0.10, 0.15, 0.20] and iteration in [400k, 500k, 600k] and found LR of 0.15 with iteration of 500k to work best.

\subsection{Quantitative Results}
\subsubsection{Sports Game Summarization Task.}
We compare our knowledge enhanced summarizer (i.e., KES) with several general text summarization models and the current state-of-the-art two-step model proposed by Huang et al.~\cite{Huang2020GeneratingSN} (i.e., SportsSUM). 
Note that the baseline models do not include the state-of-the-art pre-trained encoder-decoder language models (e.g., T5~\cite{Raffel2020ExploringTL} and BART~\cite{Lewis2020BARTDS}) due to their limitation with long text.
Tab.~\ref{table:result} shows that our model outperforms the baselines on both \textsc{K-SportsSum} and \textsc{SportsSum} datasets in terms of ROUGE scores.
Specifically, TextRank~\cite{Mihalcea2004TextRankBO} and PacSum~\cite{zheng-lapata-2019-sentence} are two typical unsupervised extractive summarization models. Roberta-Large is used as a supervised  extractive summarization model in the same way as our selector. These three models achieve limited performances due to different text styles between commentaries and news. Abs-LSTM and Abs-PGNet~\cite{See2017GetTT} are two abstractive summarization models which dispose of sports game summarization in an end2end sequence-to-sequence learning way. They outperform extractive models, because they take different text styles into account. Nevertheless, both LSTM and PGNet could not better model the long-distance dependency in the input sequence. Thanks to the appearance of transformer model~\cite{Vaswani2017AttentionIA}, an encoder-decoder architecture which makes use of self-attention mechanism to model the long-distance dependency, many pre-trained encoder-decoder language models have been proposed one after another such as BART and T5. However, they cannot be direct used for sports game summarization because the input limitations of T5 and BART are 512 tokens and 1,024 tokens, respectively.
The state-of-the-art baseline SportsSUM~\cite{Huang2020GeneratingSN} uses TextCNN selector and PGNet rewriter to achieve better results than the above models, where the selector could effectively handle the long commentaries text while the rewriter alleviates the different styles issue.
Despite its better performance, SportsSUM neglects the knowledge gap between live commentaries and sports news.
Our knowledge-enhanced summarizer uses the additional corpus to alleviate the knowledge gap, together with the advanced selector and rewriter to achieve a new state-of-the-art performance. Since \textsc{K-SportsSum} and \textsc{SportsSum} are both collected from Sina Sports Live, for each game in \textsc{SportsSum}, we also can accurately retrieve related passages and articles from the corpus, and then train our knowledge-enhanced summarizer.

% \vspace{-10ex}
\subsubsection{Ablation Study.}
We run 5 ablations, modifying various settings of our knowledge-enhanced summarizer: 
(1) remove segment embeddings in knowledge retriever; (2) remove knowledge embeddings in knowledge retriever; (3) remove both segment embeddings and knowledge embeddings; (4) replace mT5 rewriter with PGN rewriter (it is worth noting that the PGN rewriter is based on LSTM, which cannot utilize segment embeddings and knowledge embeddings); (5) replace Roberta-Large selector with TextCNN selector.

The effect of these ablations on \textsc{K-SportsSum} development set is shown in Tab.~\ref{table:ablations}. In each case, the average ROUGE score is lower than our origin knowledge-enhanced summarizer, which justifies the rationality of our model.

\subsection{Qualitative Results}
Tab.~\ref{table:case_study} shows the news sentences generated by (a) our original knowledge-enhanced summarizer, i.e., KES and (b) the variant model which removes knowledge embeddings in knowledge retriever, i.e., KES (w/o know.).
As shown, the news sentences generated by original KES are more informative than the counterpart by KES (w/o know.).
% KES successfully makes use of knowledge embedding to incorporate the additional knowledge from the external corpus.
Our knowledge-enhanced summarizer implicitly makes use of additional knowledge by fusing the knowledge embedding into the pre-trained language model (mT5 in our experiments).
Though this implicit way could help the model to generate informative sports news, we also find that this way may lead to wrong facts. As the second example shown in Tab.~\ref{table:case_study}, the generated news sentence describes the height of De Yang is 1.7 meters. However, the actual height of De Yang is 1.8 meters.
This finding implies that KES has learned the pattern of adding additional knowledge to news sentences but it is still challenging to generate correct descriptions. 

\begin{table}[t]
  \centering
  \resizebox{0.30\textwidth}{!}
  {
    \centering
    \begin{tabular}{lr}
        \hline
        Model                  & Avg. Rouge/$\triangle$ \\ \hline
        KES                    & 39.94      \\ \hline
        KES (w/o seg.)         & 39.41/-0.53      \\
        KES (w/o know.)        & 39.22/-0.72      \\
        KES (w/o seg.\&know.)  & 38.23/-1.71      \\
        KES (PGN rewriter)     & 37.02/-2.92      \\
        KES (TextCNN selector) & 37.72/-2.22      \\ \hline
        \end{tabular}
  }
  \caption{\textsc{K-SportsSum} development set ablations (seg.: segment embedding, know.: knowledge embedding).}
  \label{table:ablations}
\end{table}

\subsection{Human Study}
We conduct human studies to further evaluate the sports news generated by different methods, i.e., KES, KES (w/o know.) and SportsSUM~\cite{Huang2020GeneratingSN}. Five master students are recruited and each student evaluates 50 samples for each method.
The evaluator scores generated sports news in terms of informativeness, fluency and overall quality with a 3-point scale.

\begin{figure}[t]
\centering
\subfigure[Informativeness]{
  \includegraphics[width=0.30\linewidth]{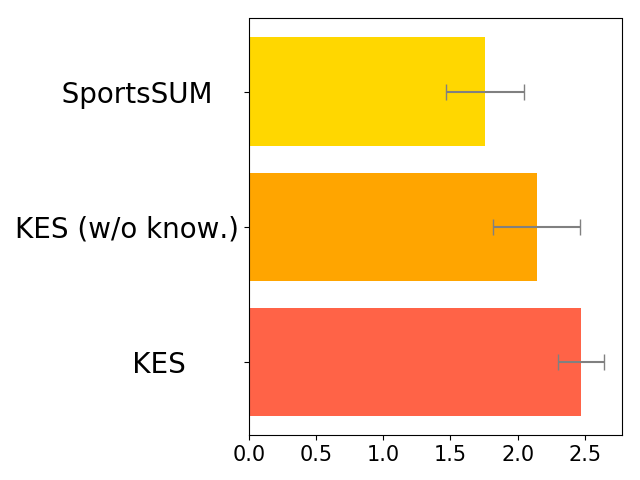}
}
\subfigure[Fluency]{
  \includegraphics[width=0.30\linewidth]{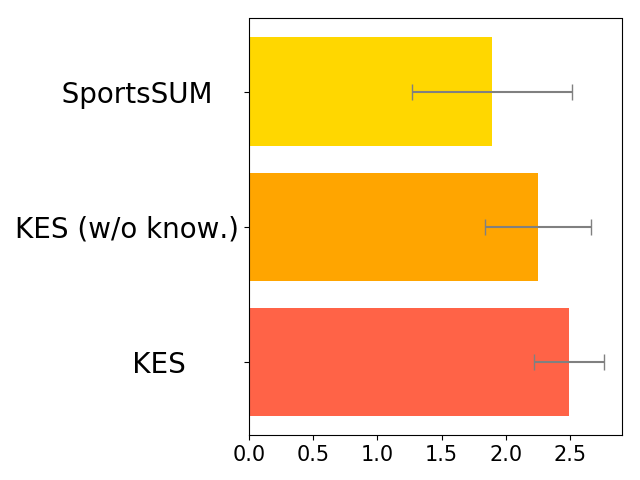}
}
\subfigure[Overall]{
  \includegraphics[width=0.30\linewidth]{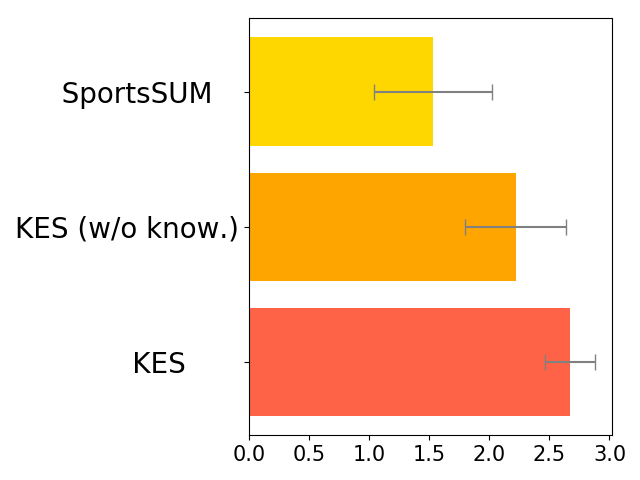}
}
% \centerline{\includegraphics[width=0.45\textwidth]{2.png}}
\caption{Results on human study.}
\label{fig:huamn_evaluation}
\end{figure}

Fig.~\ref{fig:huamn_evaluation} shows the human evaluation results. KES outperforms KES (w/o know.) and SportsSUM on all three aspects, which verifies that our original KES performances better on generating sports news. What is more, the fluency of sports news generated by KES is better than the counterpart by KES (w/o know.), which indicates taking the knowledge gap into account when generating news could also improve its fluency.

\subsection{Discussion}
We can conclude from the above experiments and analysis that sports game summarization is more challenging than traditional text summarization.
We believe the following research directions are worth following:
(1) Exploring models explicitly utilizing knowledge;
(2) Leveraging long text pre-trained model (e.g., Longformer~\cite{Beltagy2020LongformerTL} and ETC~\cite{Ainslie2020ETCEL}) to deal with sports game summarization task.

% \vspace{-10pt}
\section{Related Work}
Text Summarization aims at preserving the main information of one or multiple documents with a relatively short text~\cite{rush-etal-2015-neural,chopra-etal-2016-abstractive,Nallapati2016AbstractiveTS}.
%
% Recently, the emergence of pre-trained models~\cite{Devlin2019BERTPO,Lewis2020BARTDS,Raffel2020ExploringTL,Zhang2020PEGASUSPW} has brought text summarization into a new era.
%
Our paper focuses on sports game summarization, a challenging branch of text summarization.
Early literature mainly explores different strategies on limited-scale datasets~\cite{zhang-etal-2016-towards,Wan2016OverviewOT} to first select key commentary sentences, and then either form the sports news directly~\cite{zhang-etal-2016-towards,Zhu2016ResearchOS,Yao2017ContentSF} or relies on human-constructed templates to generate final news~\cite{Liu2016SportsNG,lv2020generate}.
Specifically, Zhang et al.~\cite{zhang-etal-2016-towards} extract different features (e.g., the number of words, keywords and stop words) of commentary sentences, and then utilize a learning to rank (LTR) model to select key commentary sentences so as to form news. Yao et al.~\cite{Yao2017ContentSF} take the description style and the importance of the described behavior into account during key commentary sentences selection. Zhu et al.~\cite{Zhu2016ResearchOS} model the sentences selection process as a sequence tagging task and deal with it using Conditional Random Field (CRF).
Lv et al.~\cite{lv2020generate} make use of Convolutional Neural Network (CNN) to select key commentary sentences and further adopt pre-defined sentence templates to generate final news.
Recently, Huang et al.~\cite{Huang2020GeneratingSN} present \textsc{SportsSum}, the first large-scale sports game summarization dataset. They also discuss a state-of-the-art two-step framework which first selects key commentary sentences, and then rewrites each selected sentence to a news sentence through seq2seq models.
Despite its great contributions, there are many noises in \textsc{SportsSum} due to its simple rule-based data cleaning process. %In contrast, \textsc{K-SportsSum} is constructed through a manual cleaning process. It's currently the highest quality and largest sports game summarization dataset.

%Unlike previous work, instead of neglecting the knowledge gap between live commentaries and sports news, \textsc{K-SportsSum} also provides a large-scale knowledge corpus.
%
%Furthermore, we propose a knowledge-enhanced summarizer that considers the knowledge from the corpus to generate sports news.

% There are also some online systems developed by industries such as Xiaoming Bot~\cite{Xu2020XiaomingbotAM} and AI football news\footnote{https://www.51zhanbao.com}.

\section{Conclusion}
In conclusion, we propose \textsc{K-SportsSum}, a large-scale human-cleaned sports game summarization benchmark. In order to narrow the knowledge gap between live commentaries and sports news, \textsc{K-SportsSum} also provides a knowledge corpus containing the information of sports teams and players. Additionally, a  knowledge-enhanced summarizer is presented to harness external knowledge for generating more informative sports news. We have conducted extensive experiments to verify the effectiveness of the proposed method on two datasets compared with current state-of-the-art baselines via quantitative analysis, qualitative analysis and human study.

\begin{acks}
Zhixu Li and Jianfeng Qu are the corresponding authors.
We would like to thank anonymous reviewers for their suggestions and comments. 
This research is supported by the National Natural Science Foundation of China (Grant No. 62072323, 61632016, 62102276), the Natural Science Foundation of Jiangsu Province (No. BK20191420, BK20210705), the Priority Academic Program Development of Jiangsu Higher Education Institutions, Suda-Toycloud Data Intelligence Joint Laboratory, and the Collaborative Innovation Center of Novel Software Technology and Industrialization.
\end{acks}

\clearpage
\bibliographystyle{ACM-Reference-Format}
\balance
\bibliography{sportssum}

% \appendix

% \section{Time Information Extraction Method}
% \label{extraction}
% \begin{figure}[t]
% \centerline{\includegraphics[width=0.5\textwidth]{re.png}}
% \caption{The regular expressions used to extract timeline information.}
% \label{fig:regular-expressions}
% \end{figure}  

% As shown in Fig.~\ref{fig:regular-expressions}, we design 22 regular expressions to extract time information for news sentences. 97.17\% of news sentences could be covered by those regular expressions if the first five words contain ``minute''. Note that, not all ``minute'' hint at the time information from news sentences. So we use those regular expressions to extract time information accurately.

\end{document}